\documentclass[10pt,twocolumn,letterpaper]{article}

\usepackage[pagenumbers]{cvpr}

\definecolor{cvprblue}{rgb}{0.21,0.49,0.74}
\usepackage[pagebackref,breaklinks,colorlinks,allcolors=cvprblue]{hyperref}
\usepackage{fontawesome5}
\usepackage[normalem]{ulem}
\usepackage{caption}
\usepackage{graphicx}
\usepackage{float}
\usepackage{siunitx}   % numeric alignment
\usepackage{xcolor}    % colors for deltas
\usepackage{colortbl}  % optional, for shaded rows
\usepackage{multirow}

\usepackage{cuted}

\useunder{\uline}{\ul}{}

\newcommand{\improv}[1]{\textcolor{green!50!black}{+#1}}
\newcommand{\dropv}[1]{\textcolor{red!70!black}{-#1}}

\title{EoS-FM: Can an Ensemble of Specialist Models \\ act as a Generalist Feature Extractor?}

\author{Pierre Adorni$^{1}$, Minh-Tan Pham$^{1}$, Stéphane May$^{2}$, Sébastien Lefèvre$^{1,3}$\\ \\
$^{1}$ IRISA, Université Bretagne Sud, UMR 6074, Vannes, France\\
$^{2}$ Centre National d’Études Spatiales (CNES), Toulouse, France \\
$^{3}$ UiT The Arctic University of Norway, Tromsø, Norway\\
{\tt\small \{pierre.adorni,minh-tan.pham,sebastien.lefevre\}@irisa.fr, stephane.may@cnes.fr}
}

\begin{document}
\maketitle

\begin{strip}
\vspace{-1cm}
    \begin{center}
        \includegraphics[width=1.0\linewidth]{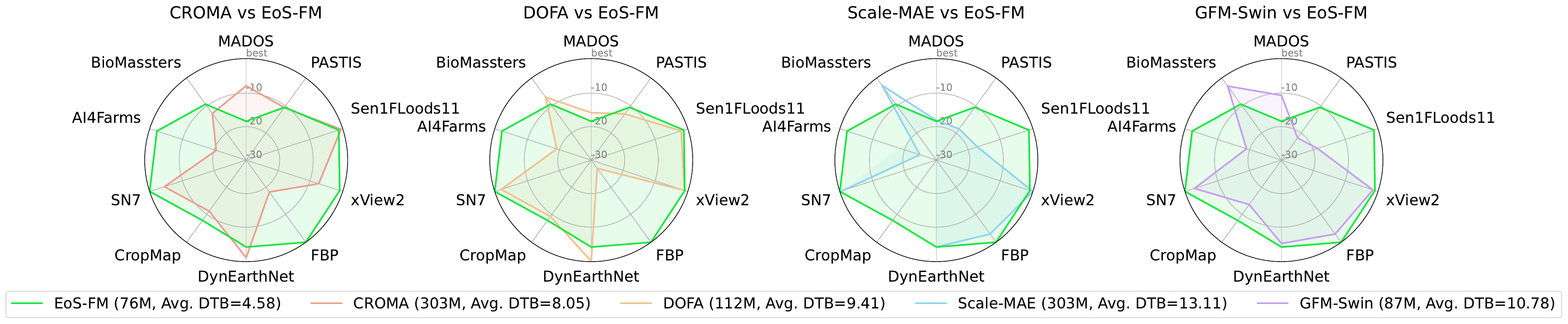}

        \captionof{figure}{The proposed EoS-FM demonstrates strong and consistent performance across 11 remote sensing tasks, emerging as the most balanced foundation model among those evaluated on the Pangaea Benchmark~\cite{marsocciPANGAEAGlobalInclusive2024}, despite having fewer parameters. For each method, we show the number of parameters and the average DTB (Distance To Best) metric (lower is better) which will be described in Sec.~\ref{sec:downstream}. }\label{fig:radarchart}

    \end{center}
\end{strip}

\vspace{-2cm}

\begin{abstract}
Recent advances in foundation models have shown great promise in domains such as natural language processing and computer vision, and similar efforts are now emerging in the Earth Observation community. These models aim to generalize across tasks with limited supervision, reducing the need for training separate models for each task. However, current strategies, which largely focus on scaling model size and dataset volume, require prohibitive computational and data resources, limiting accessibility to only a few large institutions. Moreover, this paradigm of ever-larger models stands in stark contrast with the principles of sustainable and environmentally responsible AI, as it leads to immense carbon footprints and resource inefficiency. In this work, we present a novel and efficient alternative: an Ensemble-of-Specialists framework for building Remote Sensing Foundation Models (RSFMs). Our method decomposes the training process into lightweight, task-specific ConvNeXtV2 specialists that can be frozen and reused. This modular approach offers strong advantages in efficiency, interpretability, and extensibility. Moreover, it naturally supports federated training, pruning, and continuous specialist integration, making it particularly well-suited for collaborative and resource-constrained settings. Our framework sets a new direction for building scalable and efficient RSFMs. All codes and pretrained models are available on the public repo at \href{https://github.com/pierreadorni/EoS-FM}{https://github.com/pierreadorni/EoS-FM}.
\end{abstract}
\section{Introduction}

The idea of building a foundation model for remote sensing is an appealing goal to pursue. A single model capable of handling different tasks with far fewer labels than a specialized model would be a huge benefit to the community. In recent years, the Earth Observation (EO) research community has put strong effort into developing such models, mainly through the use of upscaling techniques~\cite{chaBillionscaleFoundationModel2024, guoSkySenseMultiModalRemote2024,diasOReoleFMSuccessesChallenges2024}. This approach has already shown success in several areas of Deep Learning and Computer Vision, helping models learn more general and robust features by scaling up both model size and dataset size.

However, while upscaling improves the state of the art, the massive computational cost, energy consumption, and carbon footprint of training multi-billion-parameter models raise serious concerns about accessibility, reproducibility, and environmental impact. Their maintenance and deployment remain out of reach for many academic or operational EO actors, and their size hampers use on edge devices critical for disaster response, agriculture, and climate monitoring.

We argue that the path toward general-purpose Remote Sensing Foundation Models (RSFMs) should emphasize modularity, efficiency, and sustainability over sheer scale. To this end, we introduce a framework to train RSFMs piece by piece, following an Ensemble-of-Specialists (EoS) paradigm. Instead of a single monolithic model, we combine a diverse set of smaller encoders, each specializing in a subset of modalities or tasks, and aggregate their representations. Unlike classical ensembles, which combine full task-specific predictors to improve performance on a single task, our framework combines frozen pretrained encoders as feature extractors to construct a shared representation that generalizes across many downstream tasks. This shifts the role of ensembling from a task-level performance optimization tool to a representation-level mechanism for building general-purpose foundation models.
Our main contributions are as follows:
\begin{enumerate}
    \item A modular and scalable RSFM architecture: We introduce an Ensemble-of-Specialists (EoS) framework that enables combining multiple pre-trained encoders without retraining them jointly.
    \item A learned selection mechanism: We propose a differentiable encoder selection layer that dynamically selects the most relevant subset of encoders for a given input or task.
    \item Strong performance and efficiency: Our model achieves competitive performance compared to state-of-the-art RSFMs on the Pangaea Benchmark~\cite{marsocci2025pangaea}, while requiring significantly fewer resources to train and deploy.
    \item Built-in sustainability and adaptability: The modular design supports federated learning and pruning, making it naturally suited for low-resource or distributed training scenarios.
\end{enumerate}

\section{Related Works}

\begin{figure*}[t]
    \centering
    \includegraphics[width=\linewidth]{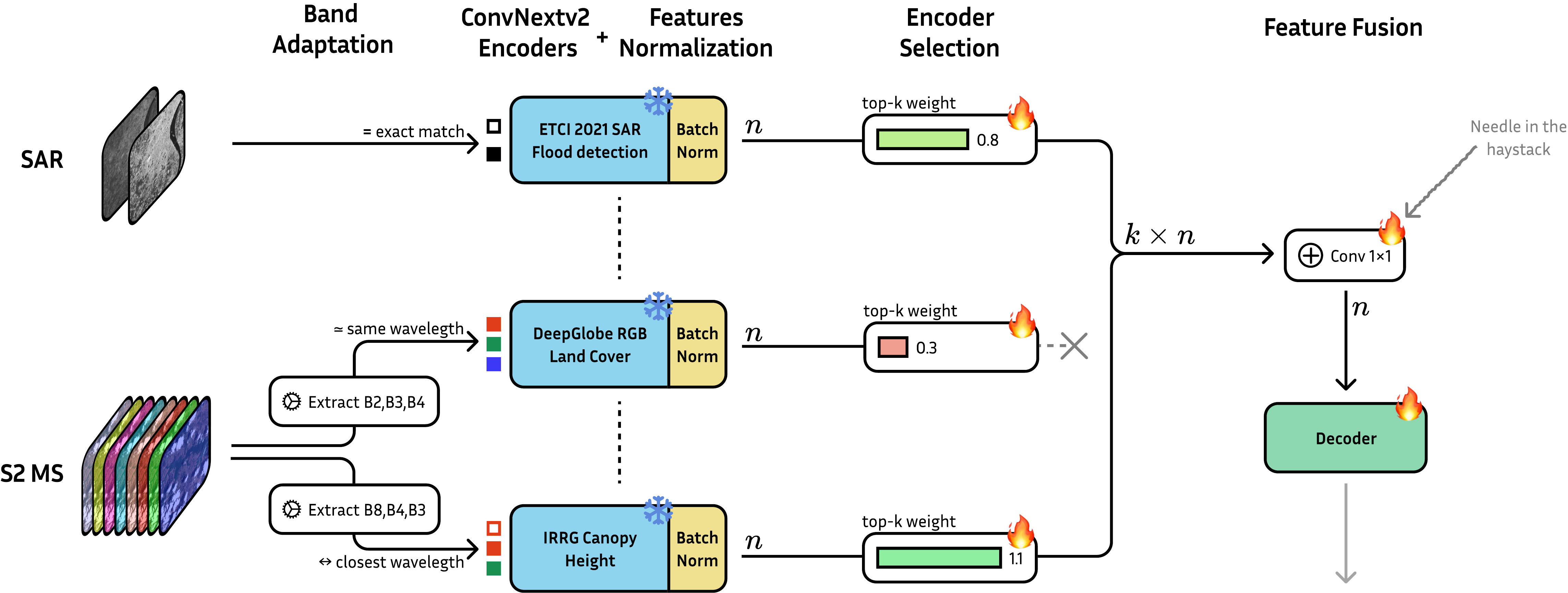}
    \caption{The EoS-FM backbone first adapts the input bands to each specialist encoder using a spectral-aware mapping, then fuses encoder outputs. A subset of $k$ encoders is selected and aggregated before being passed to the decoder.}
    \label{fig:architecture}
\end{figure*}

\textbf{Remote Sensing Foundation Models.} In Computer Vision, foundation models, typically large-scale Vision Transformers, are pretrained on vast collections of images to learn general-purpose visual features. These models can then serve as frozen backbones with task-specific decoders or be fine-tuned for a given task, requiring far less labeled data than traditional supervised approaches. The idea of building such models for EO, which involves diverse sensors, multiple spectral bands, and temporal dynamics, has gained significant traction in recent years. Over a hundred vision-only foundation models have been released since 2021~\cite{lu2025vision}, showing a clear trend toward larger architectures and pretraining datasets. Most of these efforts rely on self-supervised pretraining, as unlabeled EO imagery is abundant. Despite using datasets smaller than those in general Computer Vision (with DINOv3’s SAT493m~\cite{simeoni2025dinov3} being a notable large-scale exception), this approach has achieved strong results. We argue that incorporating supervised data, providing a more semantically meaningful training signal, could further reduce the dataset size needed while maintaining high-quality representations. Some recent works follow this direction \cite{wu2025semantic,bastaniSatlasPretrainLargeScaleDataset2023}, although they do not explore its potential for smaller and more efficient RSFMs.

\noindent\textbf{Adaptive, Modality-Agnostic RS Models.}  
Several recent works have proposed foundation models that adapt to heterogeneous inputs without requiring strict band alignment. The \textit{Panopticon} model~\cite{waldmann2025panopticon} introduces a multi-token adaptive encoder that jointly learns spatial and spectral representations across optical and SAR modalities, achieving flexible per-task adaptation. Similarly, \textit{SMARTIES}~\cite{sumbul2025smarties} learns a spectrum-aware representation space by projecting arbitrary sensor bands into a shared embedding, enabling cross-sensor and cross-mission generalization. The \textit{Copernicus-FM}~\cite{wang2025copernicus} framework leverages dynamic hypernetworks and aligned Sentinel data to support arbitrary spectral and non-spectral inputs within a single pre-trained backbone. These approaches share the goal of modality-agnostic feature extraction; however, they differ from our work in that they build a unified backbone through joint pre-training or architectural adaptation. In contrast, EoS-FM constructs a general representation by \textit{selecting independently pre-trained encoders and fusing their representations}, leveraging heterogeneity as a feature rather than consolidating it into a single monolithic model.

\noindent\textbf{Model Ensembling.} A long-established strategy for improving machine learning performance is model ensembling~\cite{dietterich2000ensemble}. In this approach, multiple models, trained independently or with slight variations, are applied to the same input and their predictions are combined, often through majority voting or averaging, to reduce bias and variance. A recent study has shown that ensembles of smaller models can even outperform single large networks in both accuracy and computational efficiency~\cite{kondratyuk2020ensembling}. More recently, feature-level ensembling has emerged as an effective alternative, where intermediate representations from multiple models are fused before classification or decoding~\cite{ullah2025hierarchical}, improving robustness and leveraging complementary information.

Unlike prior ensembling approaches, which aim to boost performance on a specific downstream task, our objective is to construct a general-purpose representation that transfers across diverse remote sensing tasks. Rather than combining full predictors trained for the same objective, we aggregate frozen pretrained encoders, each specialized on different datasets or modalities, and use their fused features as a shared backbone. In this sense, EoS-FM reframes ensembling from a task-level performance technique to a representation-level mechanism for building modular foundation models.

\noindent\textbf{Mixture of Experts.} Increasing the number of parameters generally improves model capacity but also raises inference costs. \textit{Mixture of Experts (MoE)} architectures address this issue by dividing the model into several smaller sub-networks, or experts, and activating only a subset of them for each input~\cite{jacobs1991adaptive}. A routing network learns to select which experts to use, allowing the model to maintain high representational power while reducing computational cost~\cite{shazeer2017outrageously}. Our approach shares this modular idea: the encoder is composed of multiple disjoint experts (or specialists). However, unlike traditional MoE systems, we train each expert separately on different tasks and only later train the router and feature fusion layers. This results in a sequentially-built EoS rather than a jointly-trained MoE.

\section{Proposed Method}

\subsection{Architecture Overview}

Our proposed architecture, EoS-FM (Fig.~\ref{fig:architecture}), is an ensemble of ConvNeXtV2-Atto~\cite{woo2023convnext} encoders (3.4M parameters each). We chose the ConvNeXtV2 model for its modernity and efficiency; it is known for performing well despite a small number of parameters. However, one might replace it with other backbones without any significant changes to our architecture. Each encoder is trained individually in a supervised manner on a distinct dataset and task; \eg one encoder may learn flood segmentation from SAR imagery, while another learns land use classification from multispectral data. The training procedure is detailed in Sec.~\ref{sec:training}.

\subsection{Band Adaptation}
Because the encoders in EoS-FM were trained on heterogeneous modalities and channel layouts, we include a \textit{band adaptation} step that maps the available input bands to each encoder's expected training bands. We use an adaptation layer that computes a single best mapping per encoder using remote-sensing priors.

For each required encoder band, our adaptation layer applies the following priority: (i) exact canonical band-name match, (ii) same center-wavelength match, (iii) nearest-wavelength match within the same modality, and (iv) fallback mapping when no same-modality candidate exists. This design makes adaptation more physically meaningful than purely index-based duplication/subselection and reduces spurious modality mismatches. The adapter also supports simple band expressions (e.g., $VV-VH$) by matching each operand independently and composing the resulting channel on-the-fly.

In practice, using this method, most encoders receive an input that is spectrally very close to what they have seen during training. However, when the modality shift is too large, e.g. when an encoder was pretrained on SAR data but we only have access to optical data for the task at hand, we choose to resort to less physically meaningful adaptations instead of dropping the encoder. The adaptation layer therefore prioritizes physically plausible alignment and falls back to cross-modal adaptation only as a last resort. This design is inspired by SMARTIES~\cite{sumbul2025smarties}, which projects heterogeneous sensor bands into a shared spectrum-aware space.

\begin{table*}[t]
\centering
\footnotesize
\begin{tabular}{lccccccccccc|c}
\toprule
Model & HLS & MAD & PAS & S1F & xV2 & FBP & DEN & CM & SN7 & AI4 & BM $\downarrow$ & Avg DTB $\downarrow$\\
\midrule
Scale-MAE (303M) \cite{reedScaleMAEScaleAwareMasked2023} \faSnowflake& 76.68 & 57.32 & 24.55 & 74.13 & \textbf{60.72} & 67.19 & 35.11 & 25.42 & 62.96 & 21.47 & 47.15 & 13.11 \\
Prithvi (87M) \cite{jakubikFoundationModelsGeneralist2023} \faSnowflake& 83.62 & 49.98 & 33.93 & 90.37 & 49.35 & 46.81 & 27.86 & 43.07 & 56.54 & 26.86 & 39.99 & 12.20 \\
RemoteCLIP (87M) \cite{liuRemoteCLIPVisionLanguage2024} \faSnowflake& 76.59 & 60.00 & 18.23 & 74.26 & 57.41 & {\ul 69.19} & 31.78 & 52.05 & 57.76 & 25.12 & 49.79 & 11.82 \\
SatlasNet (87M) \cite{bastaniSatlasPretrainLargeScaleDataset2023} \faSnowflake& 79.96 & 55.86 & 17.51 & 90.30 & 52.23 & 50.97 & 36.31 & 46.97 & 61.88 & 25.13 & 41.67 & 11.64 \\
S12-MAE (22M) \cite{stewart2023ssl4eo} \faSnowflake& 81.91 & 49.90 & 32.03 & 87.79 & 50.44 & 51.92 & 34.08 & 45.80 & 57.13 & 24.69 & 41.07 & 11.56 \\
SpectralGPT (105M) \cite{hongSpectralGPTSpectralRemote2024} \faSnowflake& 80.47 & 57.99 & 35.44 & 89.07 & 48.40 & 33.42 & 37.85 & 46.95 & 58.86 & 26.75 & \textbf{36.11} & 11.23 \\
S12-Data2Vec (22M) \cite{stewart2023ssl4eo} \faSnowflake& 81.91 & 44.36 & 34.32 & 88.15 & 51.36 & 48.82 & 35.90 & 54.03 & 58.23 & 24.23 & 41.91 & 11.20 \\
GFM-Swin (87M) \cite{mendietaGeospatialFoundationModels2023} \faSnowflake& 76.90 & 64.71 & 21.24 & 72.60 & 59.15 & 67.18 & 34.09 & 46.98 & 60.89 & 27.19 & 46.83 & 10.97 \\
S12-DINO (22M) \cite{stewart2023ssl4eo} \faSnowflake& 81.72 & 49.37 & 36.18 & 88.61 & 50.56 & 51.15 & 34.81 & 48.66 & 56.47 & 25.62 & 41.23 & 10.78 \\
S12-MoCo (22M) \cite{stewart2023ssl4eo} \faSnowflake& 81.58 & 51.76 & 34.49 & 89.26 & 51.59 & 53.02 & 35.44 & 48.58 & 57.64 & 25.38 & 40.21 & 10.37 \\
DOFA (112M) \cite{xiongNeuralPlasticityInspiredMultimodal2024} \faSnowflake& 80.63 & 59.58 & 30.02 & 89.37 & 59.64 & 43.18 & {\ul 39.29} & 51.33 & 61.84 & 27.07 & 42.81 & 9.41 \\
Thor Large (303M) \cite{forgaard2026thor}\faSnowflake& 79.47 & 53.73 & {\ul 39.88} & 89.55 & & 47.62 &37.29 & \textbf{60.75} &59.71 & 26.75	& & 9.25 \\
CROMA (303M) \cite{fullerCROMARemoteSensing2023} \faSnowflake& 82.42 & {\ul 67.55} & 32.32 & {\ul 90.89} & 53.27 & 51.83 & 38.29 & 49.38 & 59.28 & 25.65 & {\ul 36.81} & 8.05 \\
TerraMind Large (303M)\cite{jakubik2025terramind} \faSnowflake& 82.93 & \textbf{75.57} & \textbf{43.13} & 90.78	& & 63.38 & 37.89 & {\ul 55.04}	& 59.98	& 27.47	& & {\ul 4.65} \\
\midrule
EoS-FM (76 M) \faSnowflake& {\ul 87.01} & 61.96 & 30.30 & 89.32 & {\ul 59.79}  & \textbf{70.18} & 35.15 & 52.76 & \textbf{63.84} & {\ul 44.1} & 40.22 & \textbf{4.58} \\
EoS-FM Small (22M) \faSnowflake& \textbf{87.34} & 62.93 & 19.35 & 90.29 & 57.56 & 65.94 & 32.95 & 49.14 & {\ul 63.17} & 38.46 & 40.20 & 5.13 \\
\midrule
UNet ($\sim$8M) \cite{ronneberger2015u}\faIcon{fire}& 84.51 & 54.79 & 31.60 & \textbf{91.42} & 58.68 & 60.47 & \textbf{39.46} & 47.57 & 62.09 & \textbf{46.34} & 40.39 & 6.01 \\
ViT-B (87M) \cite{dosovitskiyImageWorth16x162020} \faIcon{fire}& 81.58 & 48.19 & 38.53 & 87.66 & 57.43 & 59.32 & 36.83 & 44.08 & 52.57 & 38.37 & 38.55 & 8.78 \\
\bottomrule
\end{tabular}
\caption{Results on the 11 downstream tasks of the Pangaea benchmark~\cite{marsocci2025pangaea}. Our method has the best or second best performance on 5 different datasets, and exhibit the best average performance, on par with TerraMind Large. (DTB~=~Distance To Best). $\downarrow$ Means lower is better.}
\label{table:results}
\end{table*}

\subsection{Encoder Selection}\label{sec:selection}

\begin{table*}[t]
\centering
\footnotesize
% \setlength{\tabcolsep}{2pt}
% \resizebox{\textwidth}{!}{
\begin{tabular}{lccccccccccc|c}
\toprule
Model & HLS & MAD & PAS & S1F & xV2 & FBP & DEN & CM & SN7 & AI4 & BM $\downarrow$ & Avg DTB $\downarrow$\\
\midrule
RemoteCLIP (87M) \faSnowflake& 69.4 & 20.57 & 17.19 & 62.22 & 53.75 & 56.23 & 34.43 & 19.86 & 43.11 & 23.85 & 53.32 & 15.95 \\
Scale-MAE (303M) \faSnowflake& 75.47 & 21.47 & 22.86 & 64.74 & \textbf{56.06} & 48.75 & 35.27 & 13.44 & 49.68 & 26.66 & 54.16 & 14.77 \\
GFM-Swin (87M) \faSnowflake& 67.23 & 28.19 & 21.47 & 62.57 & 53.45 & 55.58 & 28.16 & 27.21 & 39.48 & 32.88 & 49.3 & 14.17 \\
SatlasNet (87M) \faSnowflake& 74.79 & 29.87 & 16.76 & 83.92 & 44.07 & 37.86 & 34.64 & 29.08 & 49.78 & 13.91 & 44.38 & 13.86 \\
S12-Data2Vec (22M) \faSnowflake& 74.38 & 17.86 & 33.09 & 81.91 & 41.6 & 37.27 & 33.63 & 34.11 & 40.66 & 22.85 & 46.52 & 13.81 \\
S12-MAE (22M) \faSnowflake& 76.6 & 18.44 & 31.06 & 84.81 & 39.84 & 35.56 & 30.59 & 35.29 & 40.51 & 23.6 & 43.76 & 13.65 \\
Prithvi (87M) \faSnowflake& 77.73 & 21.24 & 33.56 & 86.28 & 35.08 & 29.98 & 32.28 & 27.71 & 36.78 & 35.04 & 41.19 & 13.48 \\
SpectralGPT (105M) \faSnowflake& 71.82 & 20.29 & 34.53 & 83.12 & 35.81 & 39.51 & 35.33 & 31.06 & 36.31 & 37.35 & \textbf{39.44} & 12.46 \\
DOFA (112M) \faSnowflake& 71.98 & 23.77 & 27.68 & 82.84 & {\ul 55.6} & 27.82 & \textbf{39.15} & 29.91 & 46.1 & 27.74 & 46.03 & 12.38 \\
S12-MoCo (22M) \faSnowflake& 73.11 & 19.47 & 32.51 & 79.58 & 41.15 & 35.57 & 32.24 & 36.54 & 49.46 & 37.97 & 44.83 & 11.82 \\
S12-DINO (22M) \faSnowflake& 75.93 & 23.47 & 36.62 & 84.95 & 41.02 & 34.63 & 32.78 & {\ul 38.44} & 41.15 & 37.91 & 42.74 & 10.78 \\
CROMA (303M) \faSnowflake& 76.44 & 32.44 & 32.8 & {\ul 87.22} & 46.54 & 37.39 & 36.08 & 36.77 & 42.15 & 38.48 & {\ul 40.25} & 8.79 \\ 
Terramind Base (86M) \faSnowflake& 77.39 & 44.06 & \textbf{39.96} & 84.43 & & 54.00 & 37.35 & 35.65 & 43.21 & 38.59 & & 5.72 \\
Thor Base (86M) \faSnowflake& 76.90	& 40.67	& {\ul 38.93} & 86.29 & & 42.80 & 35.21 & \textbf{42.23} & {\ul 55.94} & {\ul 38.90} & & {\ul 5.36} \\
\midrule
EoS-FM (76 M) \faSnowflake& {\ul 81.69} & {\ul 44.13} & 31.94 & 82.24 & 53.92 & \textbf{66.73} & \textbf{39.79} & 23.61 & \textbf{61.32} & 38.03 & 41.05 & \textbf{3.67} \\
EoS-FM Small (22M) \faSnowflake& \textbf{82.97} & \textbf{45.38} & 15.18 & 86.45 & 51.42 & {\ul 62.59} & 34.84 & 20.16 & 52.38 & 37.46 & 51.67 & 7.78 \\ \midrule
UNet ($\sim$8M)  \faIcon{fire}& 79.46 & 24.3 & 29.53 & \textbf{88.55} & 46.77 & 52.58 & 35.59 & 13.88 & 46.08 & 34.84 & 40.39 & 10.14 \\
ViT-B (87M)  \faIcon{fire}& 75.92 & 10.18 & 38.44 & 81.85 & 44.85 & 56.53 & 35.39 & 27.76 & 36.01 & \textbf{39.20} & 44.89 & 11.05 \\
\bottomrule
\end{tabular}
\caption{Results on the 11 downstream tasks of the Pangaea benchmark~\cite{marsocci2025pangaea}, using only \textbf{10\% of the training data}. Our method has the best performance on five different datasets, and exhibit the best average performance. $\downarrow$~means lower is better.
}
\label{table:results-tenpercent}
\end{table*}

One of the main advantages of our architecture lies in the \textit{modularity of the encoder} and the \textit{specialization} of its subcomponents. Different downstream tasks are likely to rely on distinct types of features, meaning that for a given task, only a subset of the encoders in the ensemble may be truly useful. In other words, the ensemble as a whole performs well across a wide variety of tasks because we can often find a subset of the specialized encoders whose learned representations partially align with the requirements of a new task. If we can identify this subset of relevant encoders, we can prune the others, yielding a much smaller model with negligible loss in performance. The central challenge, then, is to determine \textit{which encoders to keep} for each task.

To address this encoder selection problem, we introduce a lightweight \textit{selection layer} placed immediately after the feature normalization stage, inspired by Mixture-of-Experts (MoE) routing.  
Let the ensemble contain $N$ encoders $\{ E_1, E_2, \dots, E_N \}$, each producing a feature map $F_i = E_i(x)$ from an input image $x$.  
We assign to each encoder a learnable \textit{selection weight} $w_i$, initialized to $1$.  
Before fusion, each encoder’s output is scaled by its corresponding weight:
\[
\tilde{F}_i = w_i \, F_i.
\]
During \textbf{training}, we optimize these weights jointly with the task loss under an explicit sparsity regularization term,
\[
\mathcal{L}=\mathcal{L}_{\text{task}}+\lambda\,\mathcal{R}(w),
\]
where we use a \textit{topk-tail-$\ell_1$} penalty. Let $a_i = |w_i|$, and let $a_{(1)} \geq \dots \geq a_{(N)}$ denote these magnitudes sorted in descending order. The regularizer penalizes only the tail beyond rank $k$:
\[
\mathcal{R}(w)=\sum_{j=k+1}^{N} a_{(j)}.
\]
Equivalently, this is the total $\ell_1$ mass minus the top-$k$ mass:
\[
\mathcal{R}(w)=\sum_{i=1}^{N} |w_i| - \sum_{j=1}^{k} a_{(j)}.
\]
This simple objective keeps optimization fully differentiable almost everywhere while directly controlling sparsity through $k$: reducing $k$ increases the penalized tail mass and encourages only $k$ dominant encoders.

During \textbf{validation/inference}, we switch to a hard top-$k$ decision based on the learned (regularized) weights:
\[
\tilde{F}_i = 
\begin{cases}
w_i F_i, & \text{if } i \in \text{Top-}k(w), \\
0, & \text{otherwise.}
\end{cases}
\]
This train/validation split yields a stable optimization phase and a sparse, interpretable expert subset at evaluation time. This approach is similar in spirit to the experts pruning proposed by MAPEX\cite{hanna2026mapex}, which also removes modality-specific experts to obtain a lighter model at downstream time. However, while MAPEX prunes expert modules from a single jointly pre-trained Mixture-of-Experts backbone based on modality routing, our method selects among independently pre-trained encoders and performs representation-level selection rather than structural compression of a monolithic model.

Note that in the full EoS-FM configuration, $k$ is set equal to the total number of encoders ($k = N = 22$), meaning all encoders contribute to the representation and no pruning occurs at inference. The selection mechanism is only active when $k < N$: in the compact EoS-FM Small variant, we set $k = 6$, constraining the training-time selection to retain only the six most relevant encoders for each downstream task. This is discussed further in section~\ref{sec:pruning}.

\subsection{Feature Fusion}

The selected feature maps are then fused by a single $1\times1$ convolution layer, which computes linear combinations of the ensemble outputs to reduce dimensionality to a reasonable target size, by default the usual feature size of a single ConvNextv2-Atto encoder.

The fused features are then passed to a decoder to produce the final output. The ensemble functions as a foundation model: for new downstream tasks, all encoders remain frozen, and only the selection, linear fusion and decoder layers are trained. One could argue that the fusion layer is technically part of the encoder, making comparisons with fully frozen foundation models slightly unfair. However, since the fusion layer is minimal in size, we observe similar performance with or without it in practice. Keeping an explicit fusion layer is nevertheless more efficient, as it prevents the decoder from having to process excessively large feature maps.

\subsection{Training}\label{sec:training}

\begin{table*}[t]
\centering
\resizebox{0.8\textwidth}{!}{
\begin{tabular}{lllrrll}
\toprule
\textbf{Encoder Name} & \textbf{Dataset} & \textbf{Modality} & \textbf{Bands} & \textbf{Images}    & \textbf{Task} \\ 
\midrule
eurosat-s2            & EuroSat~\cite{helber2019eurosat}           & S2 MS             & 13                      & 16,200          & Scene Cls.        \\
caffe                 & Caffe~\cite{gourmelon2022calving}          & SAR               & 3                       & 13,090          & Calving Fronts Seg. \\
sen12ms-s1            & Sen12MS~\cite{schmitt2019sen12ms}          & SAR               & 3                       & 130,379         & Land Cover Seg.    \\
sen12ms-s2            & Sen12MS~\cite{schmitt2019sen12ms}          & S2 MS             & 13                      & 130,379         & Land Cover Seg.       \\
deepglobe-lcc         & DeepGlobe LCC~\cite{demir2018deepglobe}    & RGB               & 3                       & 13,000          & Land Cover Seg.       \\
bigearthnet-s1        & BigEarthNetV2~\cite{clasen2024reben}       & S1                & 3                       & 237,871         & Scene Cls.        \\
eurosat               & EuroSat~\cite{helber2019eurosat}           & RGB               & 3                       & 16,200          & Scene Cls.        \\
bigearthnet           & BigEarthNetV2~\cite{clasen2024reben}       & S2 + S1           & 14                      & 237,871         & Scene Cls.    \\
firerisk              & Firerisk~\cite{shen2023firerisk}           & RGB               & 3                       & 70,331          & Fire risk Cls.    \\
bigearthnet-rgb       & BigEarthNetV2~\cite{clasen2024reben}       & RGB               & 3                       & 237,871         & Scene Cls.        \\
sen12ms-rgb           & Sen12MS~\cite{schmitt2019sen12ms}          & RGB               & 3                       & 130,379         & Land Cover Seg.       \\
imagenet              & ImageNet~\cite{deng2009imagenet}           & RGB               & 3                       & 1,281,167       & FCMAE        \\
minifrance            & MiniFrance~\cite{castillo2022semi}         & RGB               & 3                       & 472,476         & Land Cover Seg.       \\
etci2021              & ETCI2021                                   & SAR               & 3                       & 21,600          & Flood Seg.    \\
opencanopy            & OpenCanopy~\cite{fogel2025open}            & IR-R-G            & 3                       & 66,368          & Canopy Height Reg. \\
potsdam               & Potsdam~\cite{rottensteiner2012isprs}      & IR-R-G            & 3                       & 1,200           & Land Cover Seg.       \\
cloud\_cover          & Cloud Cover~\cite{radiant_earth_sentinel2_cloud_2022} & IR-R-G & 3                       & 11,748          & Cloud Seg. \\
inria\_aerial         & Inria~\cite{maggiori2017can}               & RGB               & 3                       & 15,500          & Building Seg.       \\
landcoverai           & LandCover.ai~\cite{boguszewski2021landcover} & RGB             & 3                       & 7,470           & Land Cover Seg.       \\
levircdplus           & Levir CD+~\cite{shen2021s2looking}         & RGB               & 3                       & 510             & Building CD   \\
loveda                & LoveDA~\cite{wang2110loveda}               & RGB               & 3                       & 2,522           & Land Cover Seg.       \\
kenyacroptype         & CV4A Kenya Crop Type\cite{CV4AKenya2020}   & S2 MS               & 13                      & 4,836            & MultiTemporal Seg. \\
\bottomrule
\end{tabular}}
\caption{Summary of the datasets used to train the EoS-FM ensemble. Notations S1: Sentinel-1; S2 MS: Sentinel-2 Multispectral, SAR: Synthetic Aperture Radar, IR-R-G: Infrared-Red-Green, Cls: Classification, Seg: Segmentation, CD: Change Detection.}
\label{tab:train-datasets}
\end{table*}

We train 22 \textit{ConvNeXtV2-Atto} encoders across 17 datasets and 4 modalities. The datasets were selected from the \textit{torchgeo} library~\cite{Stewart_TorchGeo_Deep_Learning_2024}  to cover a broad range of input types, including RGB, IR, multispectral, and SAR imagery, as well as different tasks such as classification, segmentation, and change detection. The trained encoders and related data modalities are described in Table~\ref{tab:train-datasets}.

Some datasets provide multiple modalities for the same geographic samples. For instance, BigEarthNet includes both Sentinel-1 and Sentinel-2 data. In such cases, we train multiple encoders by selecting subsets of the available bands or modalities, forcing each encoder to specialize in the information contained in specific inputs. For example, we train three encoders on BigEarthNet: one using both Sentinel-1 and Sentinel-2 data (14 channels total), one using only Sentinel-1 only and one using the RGB channels extracted from the Sentinel-2 image only. In total, EoS-FM is trained on 1,085,101 unique samples. Some samples are viewed multiple times under different modalities due to the modality subsampling strategy, as previously illustrated with BigEarthNet.

Each encoder is initialized from a \textit{ConvNeXtV2-Atto} model pretrained in a self-supervised manner on ImageNet (via the \texttt{timm} library~\cite{rw2019timm}), and fine-tuned until convergence on its respective dataset. When input images are too large, we use a tiling strategy with a convenient patch size (\eg, $512 \times 512$ for MiniFrance).

\section{Experiments}

\subsection{Downstream Tasks} \label{sec:downstream}

We follow the evaluation protocol of the Pangaea Benchmark~\cite{marsocci2025pangaea}, where the EoS-FM encoder ensemble is frozen and a UperNet decoder \cite{xiao2018unified} is fine-tuned for 80 epochs with a batch size of 8. The best checkpoint is selected based on validation mIoU, and final results are reported using test mIoU for segmentation tasks and test RMSE for the single regression task, BioMassters.

The Pangaea benchmark assesses the generalization ability of foundation models, how transferable their learned features are across diverse Earth observation tasks, by counting how often each model ranks in the top two positions across benchmarks. While this metric highlights models achieving state-of-the-art (SOTA) performance on multiple datasets, it overlooks models that consistently perform close to the SOTA across all tasks. We argue that the latter quality, balanced generalization, is essential for a robust, well-rounded foundation model.

To better quantify this, we employ the \textit{Average Distance To Best} (Avg.~DTB) metric proposed in \cite{adorni2025towards}, which computes the mean absolute difference between a model’s performance and the best score achieved on each dataset. Unlike \emph{“number of top-2s”} metric reported in the Pangaea Benchmark, Avg.~DTB rewards models that perform reliably well on all tasks rather than excelling at a few while failing on others.

We compare our method to all models included in the original release of the Pangaea benchmark, as well as to two recent baselines that evaluated themselves on Pangaea in their respective works: Terramind~\cite{jakubik2025terramind} and Thor~\cite{forgaard2026thor}. The authors of Terramind and Thor did not report results on two datasets from the Pangaea benchmark (xView2 and BioMassters), following the recommendation of the Pangaea authors regarding reproducibility considerations. In our case, we were able to obtain stable results on these two datasets and therefore report the performance of EoS-FM across all tasks. The DTB metric is computed using only the available performance values for each model: over 11 tasks for all models except Terramind and Thor, for which it is computed over 9 tasks.

As shown in Table~\ref{table:results}, EoS-FM achieves the lowest Avg.~DTB (4.58) among all tested models, on par with TerraMind Large (4.65). This indicates that, despite having one-fourth the parameters of the largest tested models, EoS-FM is the most consistent across all 11 downstream tasks. TerraMind and EoS-FM are the only tested models that outperform the supervised baselines (UNet Avg. DTB 6.01).  EoS-FM ranks first on the FiveBillionPixels, HLS Burn Scars, and Spacenet 7 Change Detection dataset, and second on AI4SmallFarms and xView2, demonstrating its ability to generalize across very heterogeneous remote sensing modalities: from high-resolution optical imagery to coarse multi-temporal data. On the AI4Farms dataset in particular, it significantly surpasses all other RSFMs, being the only frozen model to approach the performance of a fully trained encoder.

Another notable finding concerns the UNet baseline. The original Pangaea authors observed that, in the full data regime, this fully supervised baseline outperformed all RSFMs on average, suggesting a gap between pretraining and task-specific adaptation. Using our new metric, this finding stands: the best frozen model of the original Pangaea benchmark (CROMA) achieves an average DTB of 8.05, which underperforms the 6.01 of Unet. However, both TerraMind and our model close this gap completely: they significantly surpasses the UNet in Avg.~DTB (4.58 and 4.65 vs. 6.01), while maintaining the benefits of frozen encoder adaptation and pretraining versatility.

In contrast, several other foundation models exhibit strong but uneven performance. For example, CROMA excels on MADOS and Sen1Floods11 but lags on FiveBillionPixels and AI4Farms. Scale-MAE achieves SOTA scores on xView2 and SpaceNet7, but remains the less performing one in terms of Avg.~DTB (13.11). SpectralGPT and the S12 family of models also show solid results on individual benchmarks but fall short in overall consistency. These fluctuations underline that many RSFMs remain specialized, not truly general-purpose: a limitation EoS-FM directly addresses through its encoder ensemble design.

\subsubsection{Label Scarcity.}
One of the central promises of foundation models is the ability to perform well with limited supervision. In Earth Observation, where labeled data is expensive and often imbalanced, particularly for rare events such as natural disasters, this ability is crucial. Following the Pangaea protocol, we train a UperNet decoder on only 10\% of the available labels per task, using stratified sampling to preserve class balance.

Results in Table~\ref{table:results-tenpercent} confirm that EoS-FM retains its advantage under severe label scarcity. It achieves the best performance on five datasets and maintains the best average DTB (3.67), outperforming all baselines. This stability under low-data regimes highlights that EoS-FM’s features are both rich and transferable, requiring fewer labeled examples to adapt effectively. Notably, EoS-FM and Thor are the only models whose Avg. DTB improves under label scarcity, reflecting smaller performance degradation rather than higher absolute scores. The smaller variant, EoS-FM Small (22M parameters), even outperforms the full ensemble on several datasets (HLS Burn Scars, MADOS, Sen1Flood11), despite its lower overall Avg. DTB (7): when labels are scarce, reducing the number of feature maps can mitigate overfitting, yielding a cleaner representation space and better generalization.

Overall, these results collectively demonstrate that EoS-FM is not only well-rounded and consistent but also highly label-efficient, establishing it as a strong candidate for a general-purpose RSFM.

\subsection{Scaling \& Pruning}\label{sec:pruning}

\begin{figure*}[t]
\begin{subfigure}{0.48\textwidth}
\centering
    \includegraphics[width=\linewidth]{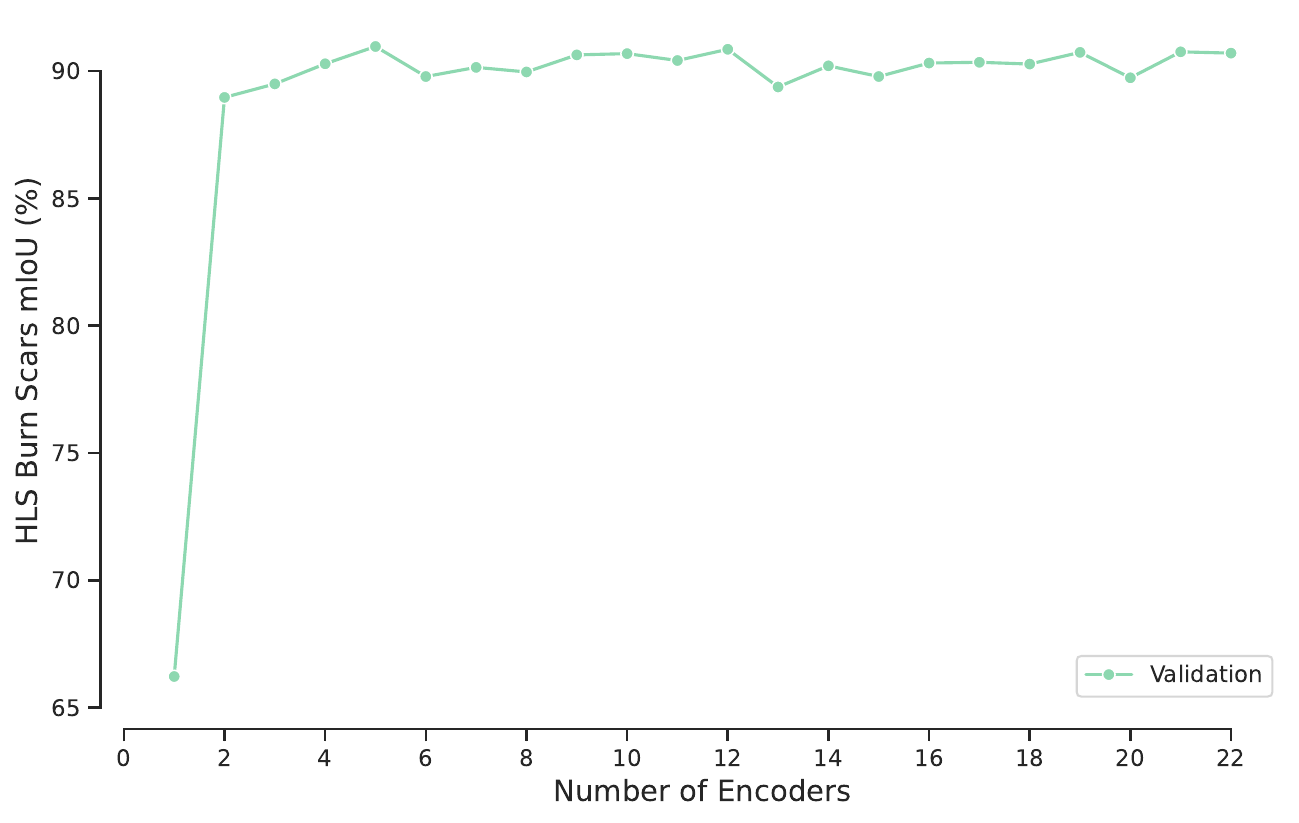}
\end{subfigure}
\hfill
\begin{subfigure}{0.48\textwidth}     
\centering
    \includegraphics[width=\linewidth]{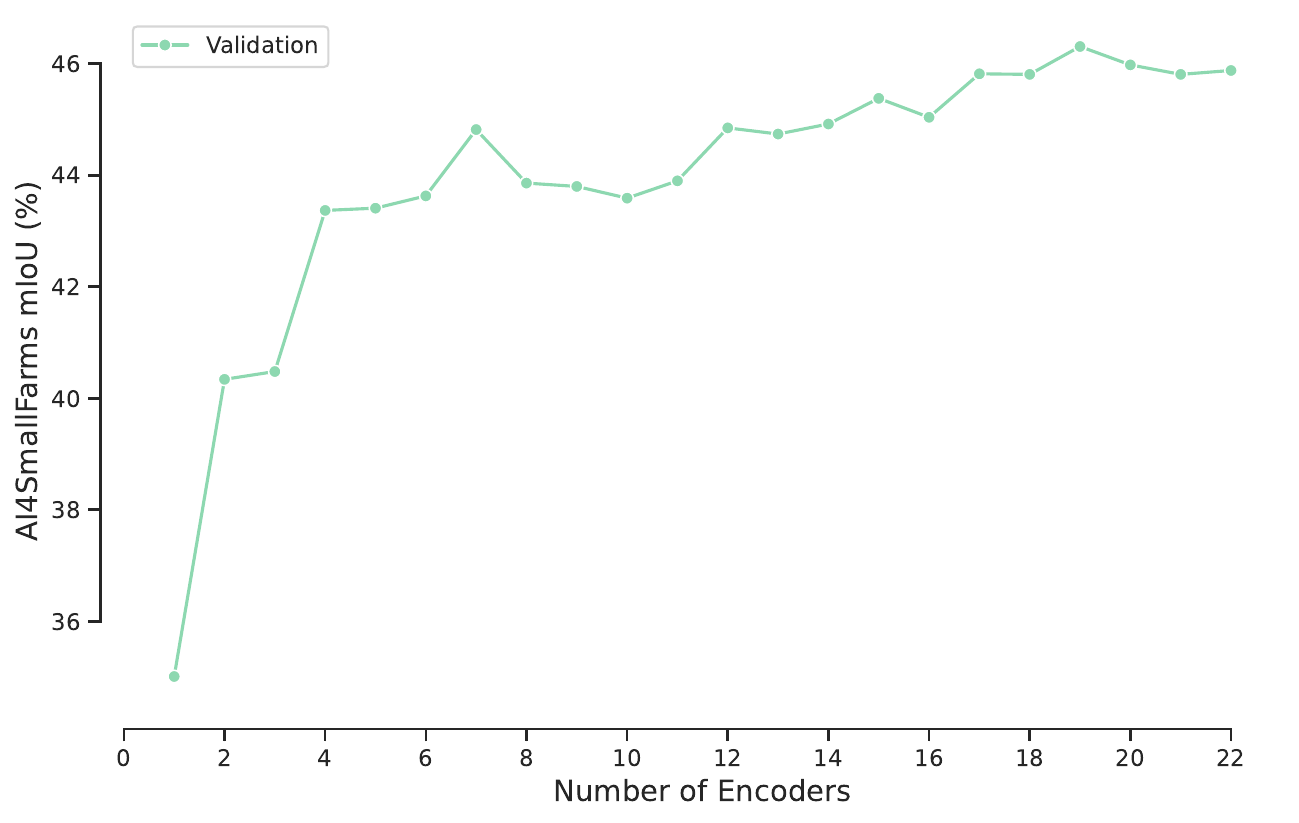}
\end{subfigure}
\caption{Ablation study: increasing the number of encoders increases the performance of the ensemble in a frozen setting. The AI4SmallFarms dataset benefits more from a larger ensemble than the HLS Burn Scars does. The validation mIoU is reported.}
\label{fig:scaling}
\end{figure*}

The results presented in the previous section were obtained with a fixed version of EoS-FM composed of 22 encoders. This configuration was chosen arbitrarily, based on our available resources and our intuition about what constitutes a diverse training set. However, the modular design of our architecture makes it straightforward to extend.

To analyze how downstream performance scales with the ensemble size, we conduct an ablation study varying the number of encoders $k$ from 1 to 22. Increasing $k$ effectively enlarges the subset of encoders selected for feature fusion until the maximum ensemble size is reached. This experiment is performed on the HLS Burn Scars and AI4SmallFarms datasets, with results presented in Fig.~\ref{fig:scaling}. Two key observations emerge:
\begin{itemize}
\item \textbf{Performance scales with ensemble size.} for HLS Burn Scars, mIoU rises quickly from 66\% with 1 encoder to 90\% with 4 encoders, and for AI4Small farms, mIoU increases more slowly from 35\% with 1 encoder to 45\% with 17 encoders. This experiment confirms that leveraging more diverse encoders improves overall performance.
\item \textbf{Performance stagnates after a few encoders.} On HLS Burn Scars, after the first 4 encoders, performance stagnates around 90\%, with slight variations due to the stochastic nature of this single-run experiment. This suggests that only a fraction of the encoders are actually able to contribute to some tasks, and reinforces the usefulness of our proposed train-time pruning strategy, described below.
\end{itemize}

The top-$k$ selection layer can serve as a mechanism for \textit{train-time pruning}, effectively controlling the model’s size during fine-tuning on a new dataset. Given a large ensemble of encoders, this layer enables the construction of task-specific lightweight variants of EoS-FM by retaining only the $k$ most relevant encoders while training the decoder. To illustrate this capability, we benchmark EoS-FM on the Pangaea datasets using $k = 6$, which reduces the total parameter count to 22M—comparable to a ResNet-50. This compact configuration, referred to as \textit{EoS-FM Small}, achieves competitive results while maintaining the efficiency benefits of a much smaller model. The corresponding performances are reported in Table~\ref{table:results} and Table~\ref{table:results-tenpercent}.

\subsubsection{Analysis of the selected encoders.}
After training the EoS-FM Small variant, we analyzed which encoders were retained or discarded by the selection layer. In many cases, the selected subsets appear intuitively relevant in a post-hoc analysis. For instance, on Sen1Floods11 — a dataset combining Sentinel-1 and Sentinel-2 imagery — the selected encoders include eurosat (S2), bigearthnet (S1+S2), sen12ms (S2), caffe (S1), etci2021 (S1), and bigearthnet (S1), all of which were pretrained on compatible modalities.

Similarly, for Five Billion Pixels, a very high-resolution (VHR) dataset, the selected encoders predominantly correspond to high-resolution pretraining sources, including eurosat (S2), landcover.ai (VHR), ImageNet, Potsdam (VHR), opencanopy (VHR), and loveda (VHR).

However, we also observe several intriguing behaviors. The encoders are not selected with equal frequency across tasks; some appear consistently more useful within the Pangaea Benchmark. Moreover, certain selections are less immediately intuitive. For example, on SpaceNet7 Change Detection: a high-resolution building construction dataset, the Inria Aerial encoder (HR building segmentation) was not selected, despite apparent task similarity, whereas the sen12ms (S1) encoder, pretrained exclusively on SAR imagery, was retained.

These observations suggest that encoder utility is not determined solely by superficial task similarity or modality alignment. A deeper investigation of these dynamics is further explored in the supplementary material.
\subsection{Feature normalization}

\begin{table*}[t]
\parbox{.50\textwidth}{
\centering
% \resizebox{\linewidth}{!}{
    \begin{tabular}{l l l l}
    \toprule
    \textbf{Dataset} &
    {\textbf{$\varnothing$}} &
    {\textbf{LayerNorm}} &
    {\textbf{BatchNorm}} \\
    \midrule
    HLS Burn Scars  & 80.52 & 84.34 {\footnotesize\improv{3.82}}  & \textbf{84.41} \footnotesize{\improv{3.89}} \\
    MADOS           & \textbf{66.06} & 64.05 \footnotesize{\dropv{2.05}}   & 62.42 \footnotesize{\dropv{3.64}}  \\
    Sen1FLoods11    & 87.95 & \textbf{89.22} \footnotesize{\improv{1.27}}  & 89.02 \footnotesize{\improv{1.07}} \\
    CropTypeMapping & 15.14 & 45.13 \footnotesize{\improv{29.99}} & \textbf{60.70} \footnotesize{\improv{45.56}} \\
    SpaceNet 7 CD   & 54.85	& 55.43 \footnotesize{\improv{0.58}}  & 54.76 \footnotesize{\dropv{0.07}} \\
    AI4SmallFarms   & 37.55 & 38.04 \footnotesize{\improv{0.49}}  & 39.56 \footnotesize{\improv{2.01}} \\
    \midrule
    \textbf{Avg.} & 57.01	& 62.70 \footnotesize{\improv{5.69}} & \textbf{65.15} \footnotesize{\improv{8.14}} \\
    \bottomrule
    \end{tabular}
% }
\centering
\caption{Effect of different feature normalization strategies on ensemble performance. val mIoU is reported for each dataset.}
\label{tab:ablation-norm}
}
\hfill
\parbox{0.40\linewidth}{
\centering
\setlength{\tabcolsep}{4pt}
% \resizebox{\linewidth}{!}{
    \begin{tabular}{lcc}
    \toprule
    Model & GFLOPs & FPS \\
    \midrule
    SSL4EO (ResNet50) & 8.2 & \textbf{2664} \\
    EoS-FM Small & \textbf{7.2} & 664 \\
    DOFA & 17.8 & 568 \\
    EoS-FM & 25.4 & 208 \\
    CROMA & 126.4 & 200 \\
    Scale-MAE & 120.0 & 168 \\
    \bottomrule
    \end{tabular}
% }
\caption{Computational efficiency comparison on RTX 5000 GPU (bs=8, $224\times224$).}
\label{tab:efficiency}
}
\end{table*}

We motivate the use of Batch Normalization over the encoder features by drawing from previous work on feature distillation~\cite{wei2205contrastive, ranzinger2410phi}, which highlights the importance of normalizing teacher features to mitigate issues caused by differences in magnitude. However, the cited approaches implement normalization differently: \cite{ranzinger2410phi} introduces a custom invertible normalization scheme to recover the original feature space, while \cite{wei2205contrastive} employs Layer Normalization. Since our pipeline does not require inverting the transformation, we opted for a simpler, non-parametric Batch Normalization layer.

To empirically validate this choice, we conducted an ablation study comparing no normalization, Layer Norm, and Batch Norm across multiple datasets (Table~\ref{tab:ablation-norm}). On average, Batch Normalization yields the best results, largely due to a substantial improvement on the CropTypeMapping dataset (+45.56 mIoU). Interestingly, the MADOS dataset shows the opposite trend: performance improves when normalization is removed altogether, suggesting that the effect of normalization may depend on the underlying feature distribution of each dataset.

\subsection{Computational Efficiency}

In addition to predictive performance, we evaluate the computational efficiency of EoS-FM and compare it with representative Remote Sensing Foundation Models (RSFMs). Table~\ref{tab:efficiency} reports the computational cost in GFLOPs and inference throughput in frames per second (FPS). All measurements are conducted on an NVIDIA RTX 5000 Ada GPU with batch size 8 and input resolution $224\times224$. GFLOPs are computed as $2\times$GMACs using the \texttt{ptflops} library, and FPS is measured during inference with frozen encoders and a UperNet decoder.

The results show that EoS-FM achieves competitive efficiency despite combining multiple pretrained encoders. The lightweight variant, EoS-FM Small, requires only 7.2 GFLOPs, making it more efficient than DOFA while maintaining strong downstream performance (Sec.~\ref{sec:downstream}). The full EoS-FM model requires significantly fewer FLOPs than large-scale foundation models such as CROMA and Scale-MAE, reducing computational cost by approximately 5$\times$ while achieving superior balanced generalization.

We note that inference throughput does not scale linearly with FLOPs, as execution efficiency also depends on hardware utilization and kernel optimization. Highly optimized architectures such as ResNet and Transformer backbones benefit from mature CUDA kernels and efficient tensor core utilization. In contrast, EoS-FM combines multiple independent ConvNextv2 encoders, which introduces additional memory and scheduling overhead despite lower total FLOPs.

\section{Discussion}

\noindent\textbf{Generalization through specialization.} 
Our results show that EoS-FM achieves strong and balanced performance across diverse tasks, suggesting an alternative path to general-purpose representations. While most foundation models rely on large-scale joint pretraining of a single backbone, EoS-FM aggregates multiple pretrained specialists. Despite being optimized for specific tasks or modalities, their fusion yields representations that generalize consistently. This indicates that general-purpose features may emerge not only from scaling individual models, but also from composing diverse specialized encoders.

\noindent\textbf{Standalone feature extraction.} 
Like other RSFMs, EoS-FM generalizes to new objectives by attaching a lightweight decoder to frozen encoders and fine-tuning only the decoder and fusion layer. However, the raw concatenation of all encoder features produces prohibitively large representations, limiting direct use as a universal embedding model. Future work could explore training a compact fusion module as a shared projection head to enable efficient standalone feature extraction.

\noindent\textbf{Federated learning.} 
The modular design naturally supports federated learning. Encoders can be trained independently on local or proprietary datasets and later integrated into the ensemble without sharing raw data. This enables collaborative and privacy-preserving scaling of Earth Observation models, aligning with sustainability goals.

\noindent\textbf{Broader impact.} 
Although introduced for remote sensing, the Ensemble-of-Specialists paradigm is domain-agnostic. Composing lightweight, task-specialized encoders and fusing their representations may offer a modular alternative to increasingly large monolithic models in any multimodal or heterogeneous setting.

\section{Conclusion}

We have introduced EoS-FM, an ensemble-based framework for building efficient and modular foundation models in remote sensing. By training lightweight specialized encoders on diverse datasets and fusing their representations, our method achieves competitive performance compared to much larger models, while remaining scalable and easily prunable. These results highlight the potential of composing FMs from smaller, domain-specialized components rather than relying solely on monolithic architectures.

Beyond performance, our approach introduces several methodological contributions. We propose a model design that is inherently modular and scalable, allowing new encoders to be added incrementally and outdated ones to be replaced without retraining the entire ensemble. Our top-k selection layer enables train-time pruning, making it possible to derive compact task-specific variants such as \textit{EoS-FM Small} from a larger ensemble. Finally, the architecture naturally supports federated learning setups, where encoders can be trained independently in distributed data environments and later integrated through the fusion module. Together, these elements illustrate a different path toward sustainable and collaborative RSFM development, grounded in flexibility, efficiency, and extensibility.

\section*{Acknowledgments}

This project was provided with computing HPC and storage resources by GENCI at IDRIS thanks to the grant 2025-AD011015819R1 on the supercomputer Jean Zay's V100 and H100 partitions. The authors would like to thank the NASA Earth Science Data Systems Program, NASA Digital Transformation AI/ML thrust, and IEEE GRSS for organizing the ETCI competition.

{
    \small
    \bibliographystyle{ieeenat_fullname}
    \bibliography{bibliography}
}

% \input{sec/7_supplementary}

\end{document}

\maketitle

\appendix
\section{Introduction}

Our main paper introduced a new framework for building Remote Sensing Foundation Models (RSFMs), along with a pretrained model, EoS-FM. We described the architecture, training, evaluation, as well as some experiments. This supplementary document provides additional experiment and analyses supporting the findings presented in the main paper. We expand on the choices made by our encoder selection layer during fine-tuning, provide more details on the training and evaluation of EoS-FM, and discuss methodological choices that were only briefly introduced in the main text. These additions are meant to give a more complete view of EoS-FM’s behavior across datasets, clarify aspects of the experimental setup, and ensure full 
reproducibility of our approach.

\section{Feature Map Normalization}
\label{sec:appendix-norm}

\begin{figure}[h]
    \centering
    \includegraphics[width=1\linewidth]{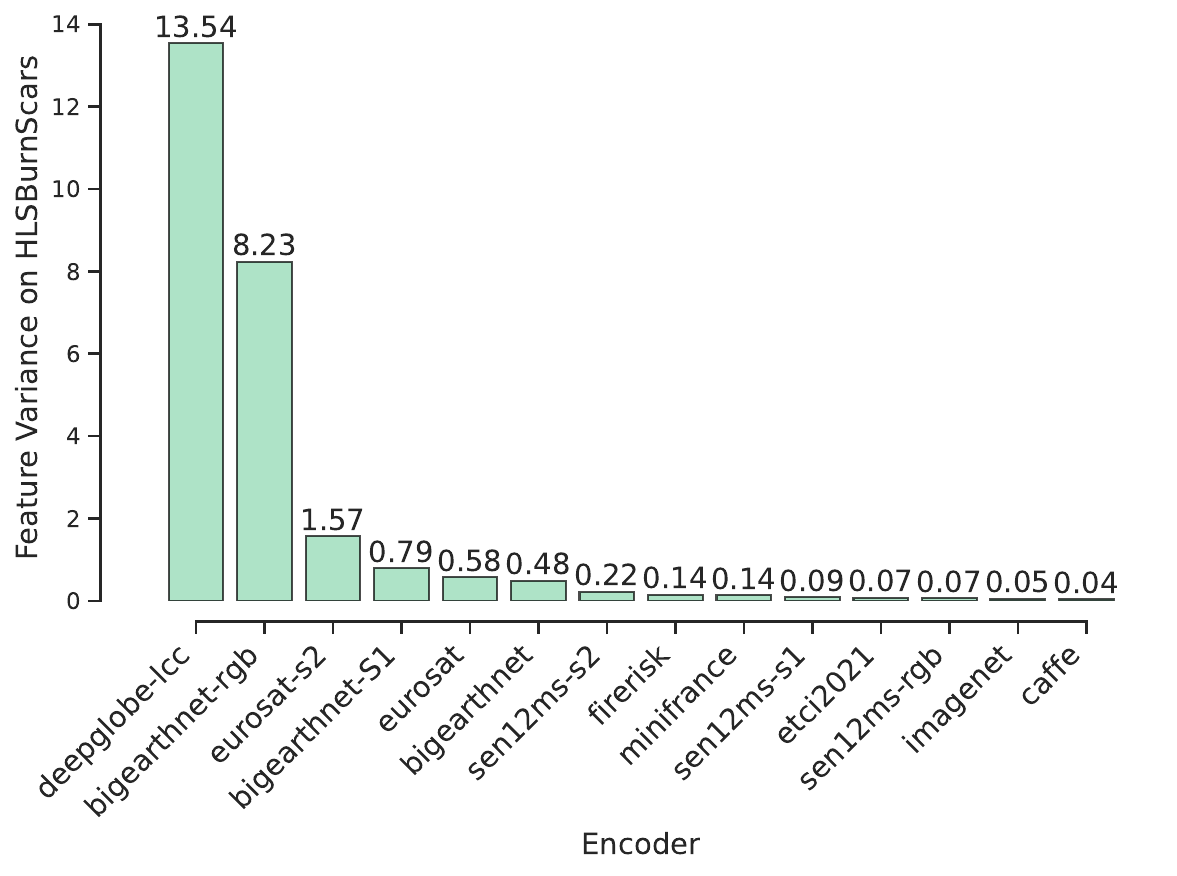}
    \caption{Variance of the feature maps computed by different encoders on the HLS Burn Scars training set. The variance changes a lot between encoders, which could create problems when training an ensemble.}
    \label{fig:feature_variance}
\end{figure}

The encoders composing our ensemble were trained on different tasks with different decoders, and therefore operate in distinct embedding spaces, producing feature maps that follow different distributions. Additionally, the semantic and modality shifts between an encoder's pretraining data distribution and the downstream data distribution is likely to result in abnormally distributed feature maps. While this is usually not a major issue when the decoder is fully trained, in our case, where the features of multiple frozen encoders are fused, it can cause the fusion layer to rely disproportionately on the \textit{loudest} encoders rather than treating all encoders equally. Figure~\ref{fig:feature_variance} illustrates this imbalance by showing the variance of feature maps across some encoders in our ensemble.

To mitigate this issue, we introduce a \textit{feature normalization stage} applied to each encoder's output before fusion. Concretely, a Batch Normalization layer is inserted after each specialist encoder and before the concatenation of their feature pyramids. This stage has no learnable parameters beyond the running statistics, and is placed immediately before the encoder selection layer and the subsequent decoder. Its role is to bring all feature maps onto a common scale, ensuring that the fusion layer allocates attention based on the semantic content of the features rather than their raw magnitude.

The motivation for this design draws from prior work on feature distillation~\cite{wei2205contrastive, ranzinger2410phi}, which highlights the importance of normalizing teacher features to mitigate magnitude imbalances. We chose Batch Normalization over alternatives such as Layer Normalization or a custom invertible scheme~\cite{ranzinger2410phi} for its simplicity and because our pipeline does not require inverting the transformation. An ablation study comparing these choices is reported in the main paper (Section~4.3), confirming that Batch Normalization yields the best average performance across Pangaea tasks.

\section{EoS-FM Training Procedure}

\begin{table}[t]
\centering
\small
\setlength{\tabcolsep}{6pt}
\begin{tabular}{lcc}
\toprule
Dataset / Encoder & Epochs & Wall-clock (h) \\
\midrule
bigearthnet & 10 & 1.74 \\
bigearthnet-rgb & 10 & 0.72 \\
bigearthnet-S1 & 10 & 0.93 \\
caffe & 100 & 2.89 \\
cloud\_cover & 20 & 0.70 \\
deepglobe-lcc & 100 & 2.68 \\
etci2021 & 100 & 1.67 \\
eurosat & 10 & 0.23 \\
eurosat-s2 & 10 & 0.05 \\
firerisk & 50 & 23.79 \\
imagenet (pretrained) & pretrained & 0.00 \\
inria\_aerial & 20 & 0.44 \\
kenya-croptype & 200 & 3.38 \\
landcoverai & 100 & 1.64 \\
levircdplus & 100 & 0.58 \\
loveda & 100 & 2.31 \\
minifrance & 10 & 19.13 \\
opencanopy & 20 & 3.03 \\
potsdam & 100 & 0.37 \\
sen12ms-rgb & 20 & 4.81 \\
sen12ms-s1 & 10 & 5.09 \\
sen12ms-s2 & 10 & 3.22 \\
\midrule
\textbf{Total} &  & \textbf{79.38} \\
\bottomrule
\end{tabular}
\caption{Dataset-wise pretraining budget for the 22 encoders in EoS-FM. The ImageNet row corresponds to pretrained initialization and therefore has no additional training time in our training pipeline.}
\label{tab:training_budget}
\end{table}

Owing to its ensemble architecture, the training of EoS-FM is inherently flexible. Specialists can be trained sequentially on a single GPU (at the cost of longer wall-clock time but with minimal parallel compute requirements) or concurrently across multiple nodes to reduce the overall schedule, with any intermediate strategy being equally valid depending on available hardware. This contrasts with monolithic foundation models, whose training is tightly coupled and difficult to distribute across heterogeneous compute allocations. In our experiments, the 22 specialists were trained independently, making it straightforward to resume, replace, or add specialists without retraining the entire ensemble.

Each specialist consists of a ConvNeXtV2-Atto backbone (~3.7M parameters), initialized from ImageNet-1K pretrained weights, paired with a deliberately lightweight task-specific head. Dense prediction tasks (semantic segmentation, change detection, pixelwise regression) use a UperNet decoder with only 16 internal channels and a single-layer segmentation head, also 16 channels wide, for a total head size of roughly 100k parameters — less than 3\% of the full specialist. Classification and multi-label classification tasks use a simple identity decoder followed by a single linear layer. This intentional bottleneck at the head level is a deliberate design choice: by constraining the head's capacity, we force the encoder to carry as much representational work as possible, which we hypothesize leads to richer, more transferable features. In contrast, a large decoder could compensate for a weaker encoder, producing good pretraining metrics while encoding little reusable structure.

All specialists are trained with the AdamW optimizer at a constant learning rate of $10^{-4}$, using a Dice loss for segmentation tasks and cross-entropy for binary tasks (BCE for multi-label classification), with bfloat16 mixed precision on a single GPU. Training runs for 10 to 200 epochs depending on the dataset size. Once training is complete, encoder weights are extracted from each Lightning checkpoint and assembled into a single \texttt{.pth} file that packages all specialists.

We emphasize that our objective is to demonstrate that \textit{an} ensemble of small, task-specific frozen encoders can serve as an effective generalist feature extractor, not that \textit{this particular set} of encoders is optimal. The choice of pretraining datasets was guided primarily by availability in the TorchGeo library, which facilitated rapid implementation. As we discuss in Appendix~\ref{sec:appendix-selection}, there is likely room for improvement through more principled encoder selection or by expanding the pretraining dataset pool.

\paragraph{Dataset-wise training budget.} Table~\ref{tab:training_budget} reports the wall-clock training time and number of epochs for each of the 22 encoders used in the final ensemble (including the ImageNet pretrained ConvNeXtV2-Atto encoder). The total wall-clock training time is 79.38 hours on a single \textit{Nvidia H100 80Gb} GPU.

\section{Selected Encoders Analysis}\label{sec:appendix-selection}

\begin{figure*}[t]
\parbox{0.48\linewidth}{
    \centering
    \includegraphics[width=\linewidth]{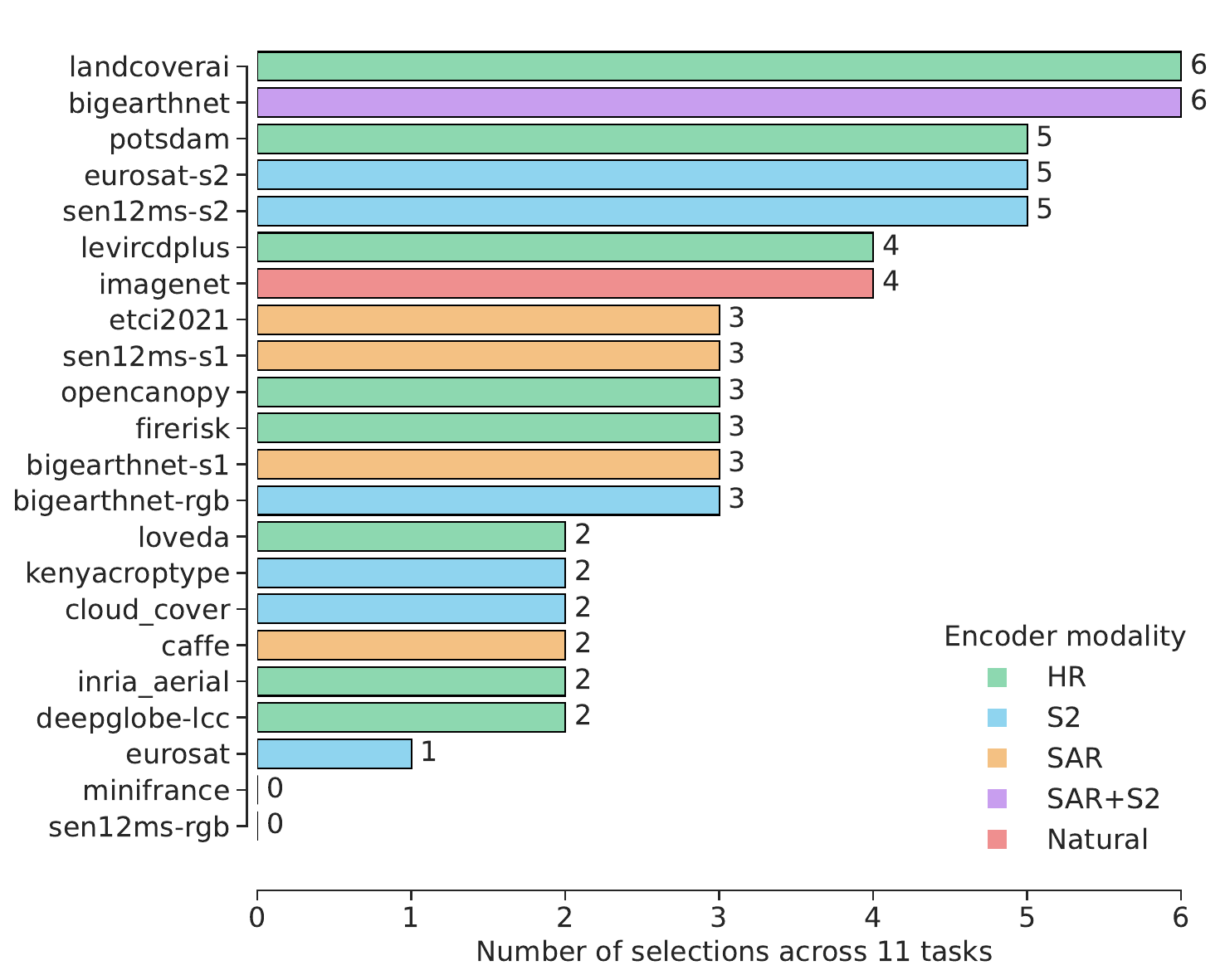}
    \caption{\textbf{Encoder selection frequency.} Some encoders are selected 6 times across Pangaea (landcoverai, bigearthnet), whereas some other are never selected (sen12ms-rgb, minifrance).}
    \label{fig:encoders-num-selections}
}
\hfill
\parbox{0.48\linewidth}{
    \vspace{0.95cm}
    \centering
    \includegraphics[width=\linewidth]{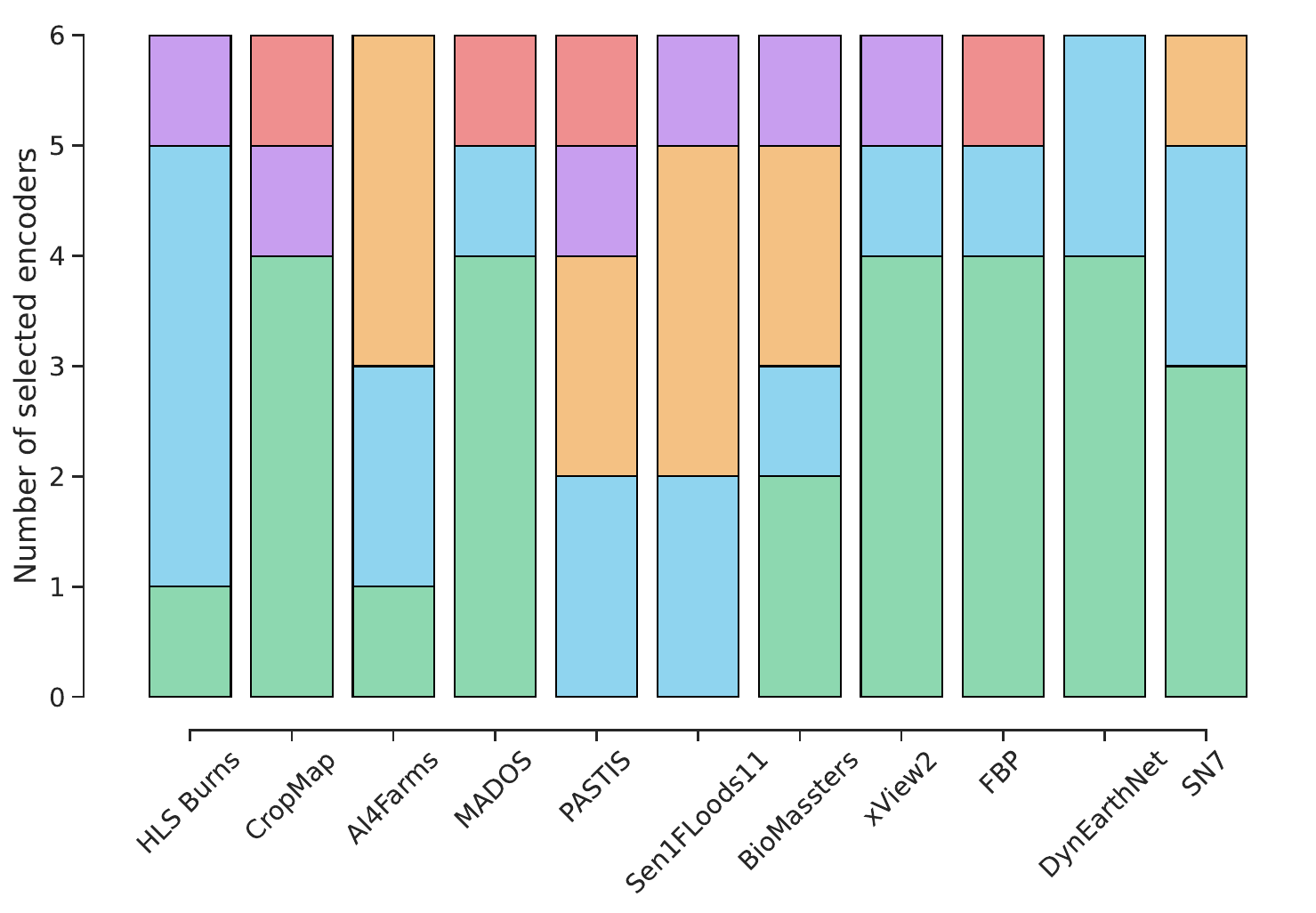}
    \caption{\textbf{Modality distribution of selected encoders for each Pangaea task.} For the majority of the tasks, the selected encoders are close modality-wise, although some inconsistent selection is observed for a minority of datasets.}
    \label{fig:encoders-selected-per-task}
}
\end{figure*}

\subsubsection{Overall selection distribution.} We analyze the encoder selections made by the selection layer during the last evaluation run of EoS-FM Small, which retains 6 encoders per downstream task. The bar plot in Figure~\ref{fig:encoders-num-selections} reports the selection frequency of each encoder. We observe substantial variability across encoders, suggesting that the representational quality of certain feature spaces is consistently superior regardless of the target application. Among the most frequently selected encoders, LandCover.ai and BigEarthNet (S1+S2) are each retained 6 times across the 11 tasks, followed by Potsdam, Eurosat (S2), and Sen12MS (S2) with 5 selections each. At the other end of the spectrum, Sen12MS (RGB) and Minifrance are never selected, and Eurosat (RGB) is retained only once.

SAR encoders are selected at most 3 times in total, which is consistent with the fact that only 4 Pangaea tasks include SAR imagery: Sen12MS, PASTIS, BioMassters, and CropTypeMapping. Nonetheless, several SAR encoders are unexpectedly selected for the AI4SmallFarms dataset, which contains only optical imagery. We investigate this observation further in Section~\ref{sec:ai4-sar}.

\subsubsection{Selected encoders per task.} Figure~\ref{fig:encoders-selected-per-task} shows the modality distribution of selected encoders per task. The selection layer behaves coherently on most datasets. For example, on HLSBurnScars (S2), the selection retains 1 SAR+S2 encoder, 4 S2 encoders, and 1 HR encoder. However, several inconsistent selections are also observed: AI4SmallFarms has 3 SAR encoders selected despite containing only optical imagery, and both MADOS and CropTypeMapping have 4 HR encoders selected despite being Sentinel-2 datasets.

To better quantify this modality misalignment, we define a \textit{Modality Coherence} score as the average, over the 6 selected encoders, of a per-encoder compatibility score. This per-encoder score reflects how well an encoder's pretraining modality matches the input modality of the downstream task. It is assigned according to the following rules, listed from most to least compatible in Table~\ref{tab:modality-alignment}.

\begin{table*}[h]
\begin{center}

\setlength{\tabcolsep}{10pt}
\begin{tabular}{lc}
\toprule
Case & Score \\
\midrule
Encoder modality matches the task modality exactly & 1.0 \\
\quad (e.g., S2 encoder for an S2 task, or S1 encoder for an S1+S2 task) & \\ \\
ImageNet-pretrained encoder & 1.0 \\ 
\quad (modality-agnostic, treated as universal) & \\ \\
Encoder modality is a strict superset of the task modality & 0.7 \\
\quad (e.g., S1+S2 encoder for an S2 task; band adaptation fills missing inputs) & \\ \\
Both encoder and task are optical, but from different sensor types & 0.2 \\
\quad (e.g., HR encoder for an S2 task, or vice versa) & \\ \\
Modalities are incompatible & 0.0 \\
\quad (e.g., SAR encoder for an optical task) & \\
\bottomrule
\end{tabular}
\end{center}
\caption{Details on the computation of the modality alignement score.}\label{tab:modality-alignment}
\end{table*}

\begin{figure}
    \centering
    \includegraphics[width=\linewidth]{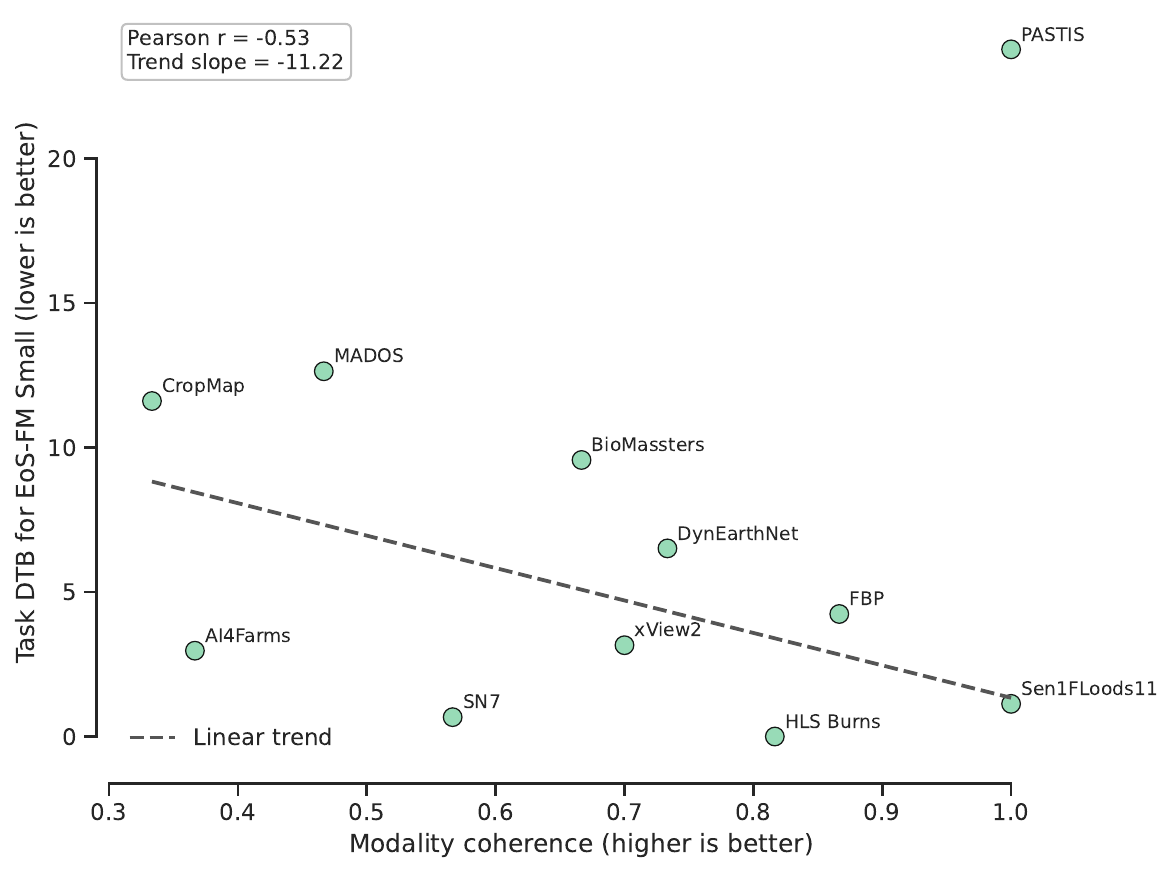}
    \caption{\textbf{Correlation between DTB metric and modality coherence.} We observe a weak correlation (r=0.53) between high modality coherence and good performance. PASTIS is left out of the trend computation as EoS-FM Small is particularly bad on this dataset and, as such, it is an outlier.}
    \label{fig:modality-coherence}
\end{figure}

We plot in figure~\ref{fig:modality-coherence} the relationship between Task \textit{Distance To Best} (DTB) and our modality coherence score for the 11 tasks of Pangaea, using the test metrics of EoS-FM Small on these datasets. When we set aside the PASTIS dataset, on which EoS-FM Small performs particularly badly, we observe a weak correlation (r=0.53) between high performance and high modality coherence.

\subsection{On AI4SmallFarms and SAR encoders} \label{sec:ai4-sar}

As noted above, some datasets exhibit very low modality coherence scores. For AI4SmallFarms (S2), the selection layer retains the following encoders: LevirCD+ (HR CD), Caffe (S1), BEN-RGB (S2), Sen12MS-S1 (S1), BigEarthNet (S1), and KenyaCropType (S2). While BEN-RGB and KenyaCropType are modality-appropriate choices, the remaining encoders are predominantly SAR-based, which is unexpected given that AI4SmallFarms provides only 4 Sentinel-2 spectral bands. To investigate this behavior, we conduct an ablation study in which the automatically selected subset is replaced by a manually curated set of modality-relevant encoders: BEN-RGB (S2), KenyaCropType (S2), ImageNet, Eurosat (S2), LandCoverAI (HR), and LevirCD+ (HR). This curated set prioritizes S2 specialists and complements them with optical encoders, without any SAR-based specialist.

We apply the same approach to MADOS (S2), which over-relies on HR encoders despite being a Sentinel-2 dataset. The curated encoder set for MADOS consists of: Eurosat (S2), Sen12MS (S2), BigEarthNet (RGB), Cloud Cover (S2), Sen12MS (RGB), and ImageNet.

Let $S_1$ denote the encoder subset selected by the selection layer, and $S_2$ the manually curated subset. Table~\ref{tab:ai4-mados-ablation} reports the training and validation mIoU for both subsets on AI4SmallFarms and MADOS.

\begin{table*}[h]
    \centering
    \setlength{\tabcolsep}{8pt}
    \begin{tabular}{lcccc}
        \toprule
        & \multicolumn{2}{c}{AI4SmallFarms} & \multicolumn{2}{c}{MADOS} \\
        \cmidrule(lr){2-3} \cmidrule(lr){4-5}
        Encoder subset & Train mIoU & Val mIoU & Train mIoU & Val mIoU \\
        \midrule
        $S_1$ (selection layer) & 44.30 & \textbf{46.21} & 86.01 & \textbf{67.67} \\
        $S_2$ (curated)         & \textbf{45.46} & 39.77 & \textbf{92.20} & 59.72 \\
        \bottomrule
    \end{tabular}
    \caption{Comparison of encoder subsets automatically selected by the selection layer ($S_1$) and manually curated ($S_2$), on AI4SmallFarms and MADOS. Bold values indicate the best result per column.}
    \label{tab:ai4-mados-ablation}
\end{table*}

The curated subset $S_2$ achieves higher training mIoU on both datasets, likely because, with the selection fixed, the decoder can be optimized without simultaneously training the selection layer. Nevertheless, $S_2$ yields significantly lower validation mIoU, indicating that the automatically selected encoders generalize better despite their apparent modality mismatch. These results suggest that the fusion layer either learns to exploit features from modality-misaligned encoders in a way that is not immediately interpretable, or effectively discards their contributions at inference time. In the latter case, the selection layer may have simply filled the remaining slots of its fixed budget of 6 encoders with whatever was available, without those encoders meaningfully influencing the decoder output.

\section{Pangaea Testing Procedure}

The Pangaea Bench codebase was used to perform the evaluation. The standard protocol, as established by the original authors, proceeds as follows:
\begin{enumerate}
    \item \textbf{Decoder.} A UperNet decoder with an internal width of 512 is attached to the encoder, operating on 4 feature levels. Since UperNet expects a feature pyramid with progressively decreasing spatial resolution, spatial pooling is applied to the feature maps of vision transformer encoders to emulate this pyramid structure. The number of decoder parameters varies with the encoder's latent dimensionality, but typically amounts to approximately 30M. For multi-temporal datasets, each time step is encoded independently to produce a sequence of feature pyramids, which are then aggregated along the temporal dimension using a Lightweight Temporal Attention Encoder (LTAE)~\cite{garnot2020ltae}: a multi-head self-attention module operating per spatial location across time steps, conditioned on sinusoidal positional encodings of the acquisition dates, whose output is further processed by a small MLP before being passed to UperNet.
    \item \textbf{Dataset.} Each input sample is randomly cropped or resized to the encoder's expected input resolution. Only the spectral bands accepted by the encoder are retained; the remaining ones are discarded. The selected bands are then normalized by their respective mean and standard deviation, and any band required by the encoder but absent from the input is zero-padded. In the case of EoS-FM, band filtering and padding are bypassed, as the model handles band adaptation internally. Similarly, input resizing is disabled: we set the encoder input size to match the native image resolution of each dataset, as the ensemble was trained across a variety of resolutions and image sizes and is therefore expected to generalize to arbitrary input dimensions.
    \item \textbf{Training.} The encoder is frozen and the UperNet decoder is trained for 80 epochs using the AdamW optimizer, with a learning rate of $10^{-4}$ for the first 60\% of training, $10^{-5}$ for the next 30\%, and $10^{-6}$ for the final 10\%. The default loss function is DICE, except on xView2, where the original authors substituted Cross-Entropy due to observed training instabilities. In our experiments, we found that EoS-FM also fails to converge reliably under the Dice loss on four datasets: xView2, MADOS, Five Billion Pixels, and CropTypeMapping, and therefore adopted Cross-Entropy for those cases as well. While this constitutes a minor deviation from the standard evaluation protocol, we argue that Cross-Entropy is unlikely to yield systematically higher performance than Dice on these tasks, and thus has a negligible effect on the comparability of our results with the baseline models.
\end{enumerate}

To obtain the results in Table~1, we perform 3 independent evaluations on each dataset with a random seed, and report the average.

{
    \small
    \bibliographystyle{ieeenat_fullname}
    \bibliography{bibliography}
}